\documentclass[runningheads]{llncs}
\usepackage{graphicx}
\usepackage{tikz}
\usepackage{comment}
\usepackage{amsmath,amssymb} % define this before the line numbering.
\usepackage{color}
\usepackage{booktabs}
\usepackage{multirow}
\usepackage{wrapfig}
\usepackage{graphicx}
\usepackage{subfigure}

\definecolor{citecolor}{RGB}{65,105,225}
\usepackage[pagebackref=true,breaklinks=true,letterpaper=true,colorlinks,
citecolor=citecolor,bookmarks=true]{hyperref}

\usepackage[accsupp]{axessibility}  % Improves PDF readability for those with disabilities.
\usepackage{url}
\usepackage[font={small}]{caption}

\usepackage{xcolor}

\begin{document}

\pagestyle{headings}
\mainmatter
\def\ECCVSubNumber{1687}  % Insert your submission number here

\title{Fast-Vid2Vid: Spatial-Temporal Compression for Video-to-Video Synthesis} % Replace with your title

\begin{comment}
\titlerunning{ECCV-22 submission ID \ECCVSubNumber} 
\authorrunning{ECCV-22 submission ID \ECCVSubNumber} 
\author{Anonymous ECCV submission}
\institute{Paper ID \ECCVSubNumber}
\end{comment}
%******************

%\begin{comment}
\titlerunning{Fast Vid2Vid}

\author{Long Zhuo\inst{1}
\and Guangcong Wang\inst{2}
\and Shikai Li\inst{3}
\and Wayne Wu\inst{1,3}
\and Ziwei Liu\inst{2}\thanks{Corresponding author}
}
\authorrunning{L. Zhuo et al.}
\institute{Shanghai AI Laboratory \and S-Lab, Nanyang Technological University \and SenseTime Research \\
\email{ zhuolong@pjlab.org.cn\quad \{guangcong.wang,  ziwei.liu\}@ntu.edu.sg \quad lishikai@sensetime.com \quad wuwenyan0503@gmail.com}}

%\end{comment}
%******************

\maketitle
\begin{figure}
  \vspace{-20pt}
    \centering
  \includegraphics[width=0.95\linewidth]{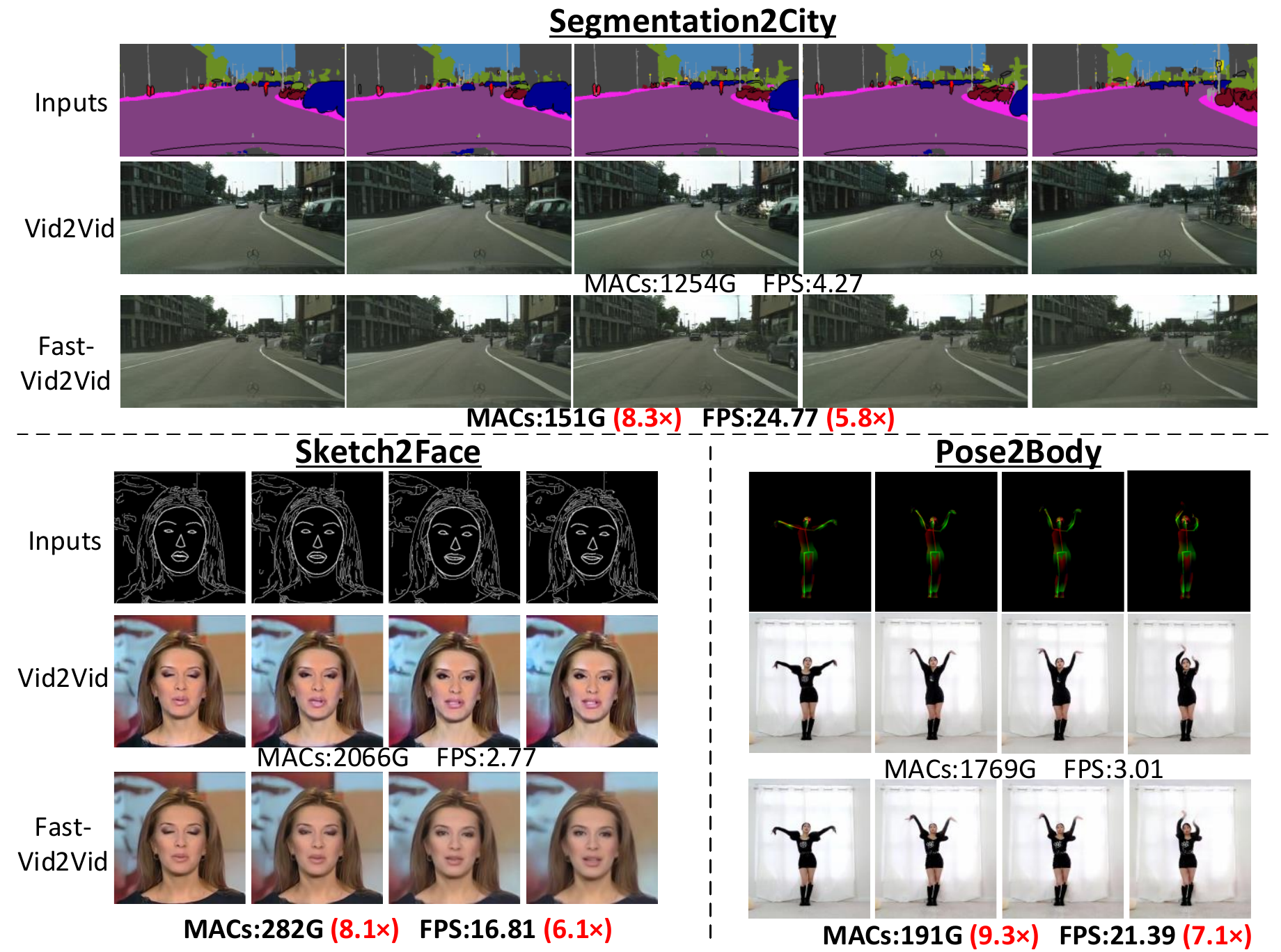}
  \vspace{-10pt}
  \caption{\textbf{Fast-Vid2Vid.} Our proposed Fast-Vid2Vid accelerates the video-to-video synthesis and generates photo-realistic videos more efficiently compared to the original vid2vid model. On standard benchmarks, Fast-Vid2Vid achieves \textcolor{orange}{\textbf{16.81-24.77 FPS}} and saves \textcolor{orange}{\textbf{8.1-9.3$\times$}} computational cost in the tasks of Sketch2Face, Segmentation2City and Pose2Body.}
  \label{fig:teaser}
\end{figure}
\vspace{-30pt}

\begin{abstract}
Video-to-Video synthesis (Vid2Vid) has achieved remarkable results in generating a photo-realistic video from a sequence of semantic maps. However, this pipeline suffers from high computational cost and long inference latency, which largely depends on two essential factors: 1) network architecture parameters, 2) sequential data stream. Recently, the parameters of image-based generative models have been significantly compressed via more efficient network architectures. Nevertheless, existing methods mainly focus on slimming network architectures and ignore the size of the sequential data stream. Moreover, due to the lack of temporal coherence, image-based compression is not sufficient for the compression of the video task. In this paper, we present a spatial-temporal compression framework, \textbf{Fast-Vid2Vid}, which focuses on data aspects of generative models. It makes the first attempt at time dimension to reduce computational resources and accelerate inference. Specifically, we compress the input data stream spatially and reduce the temporal redundancy. After the proposed spatial-temporal knowledge distillation, our model can synthesize key-frames using the low-resolution data stream. Finally, Fast-Vid2Vid interpolates intermediate frames by motion compensation with slight latency.
On standard benchmarks, Fast-Vid2Vid achieves around real-time performance as 20 FPS and saves around $8\times$ computational cost on a single V100 GPU. Code and models are publicly available\footnote{Project page: \url{https://fast-vid2vid.github.io/} \\
Code and models: \url{https://github.com/fast-vid2vid/fast-vid2vid}}.


\end{abstract}

\section{Introduction}
Video-to-video synthesis (vid2vid)~\cite{wang2018video} targets at synthesizing a photo-realistic video given a sequence of semantic maps as input. A wide range of applications are derived from this task, such as face-talking video generation $\quad$ (Sketch2Face)~\cite{wang2018video,wang2019few}, driving video generation (Segmentation2City)~\cite{wang2018video,wang2019few} and human pose transferring generation (Pose2Body)~\cite{chan2019everybody,kappel2021high,zhou2019dance}. With the advance of Generative Adversarial Networks (GANs)~\cite{gan2014nips}, vid2vid models~\cite{wang2018video,wang2019few} have made significant progress in video quality. 
However, these approaches need large-scale computational resources to yield the results, and they are computationally prohibitive and environmentally unfriendly. 
For example, the standard vid2vid~\cite{wang2018video} consumes $2066$ G MACs for generating each frame, which is $500\times$ more than ResNet-50~\cite{he2016deep}. Recent studies demonstrate that lots of recognition compression approaches have been successfully extended to image-based GAN compression methods~\cite{aguinaldo2019compressing,chen2020distilling,fu2020autogan,jin2021teachers,liu2021content,li2020gan}. Can we directly employ these existing image-based GAN compression methods to achieve promising vid2vid compression models?

In the literature, image-based GAN compression methods can be roughly categorized into three groups, including knowledge distillation~\cite{aguinaldo2019compressing,chen2020distilling,jin2021teachers,liu2021content,belousov2021mobilestylegan}, network pruning~\cite{liu2021content,wang2020gan}, and neural architecture search (NAS)~\cite{li2020gan,gong2019autogan,fu2020autogan,gao2020adversarialnas,lin2021anycost}.
They focus on obtaining a compact network by cutting the network architecture parameters of the original network. However, the input data, another factor that significantly affects the inference speed of a deep neural network, has been ignored by the existing GAN compression methods. Moreover, since they are image-based synthesis tasks, they do not consider redundant temporal information hidden in neighbor frames of a video. Therefore, directly applying image-based compression models to vid2vid synthesis is difficult to achieve the desired results. 
In this work, we aim to compress the input data stream while maintaining the well-designed network parameters, and generate the photo-realistic results for vid2vid synthesis. Furthermore, we make an initial attempt at removing temporal redundancy to accelerate vid2vid model.

There are three critical challenges for vid2vid compression. First, the typical vid2vid model~\cite{wang2018video} consists of several encoder-decoders to capture both spatial and temporal features. It is difficult to reduce the parameters from such a complicated structure due to the intricate connections between these encoders and decoders. 
Second, it is a challenge to compress the input data stream and achieve decent performance for GAN generation since the perceptual fields of GAN are much more erratic than image recognition. Third, transferring knowledge from a teacher model to a student model temporally is challenging to align with the spatial knowledge distillation as the temporal knowledge is implicitly hidden within adjacent frames and is more difficult to capture than the spatial knowledge.

\begin{figure*}[t]
  \centering
  \includegraphics[width=\linewidth]{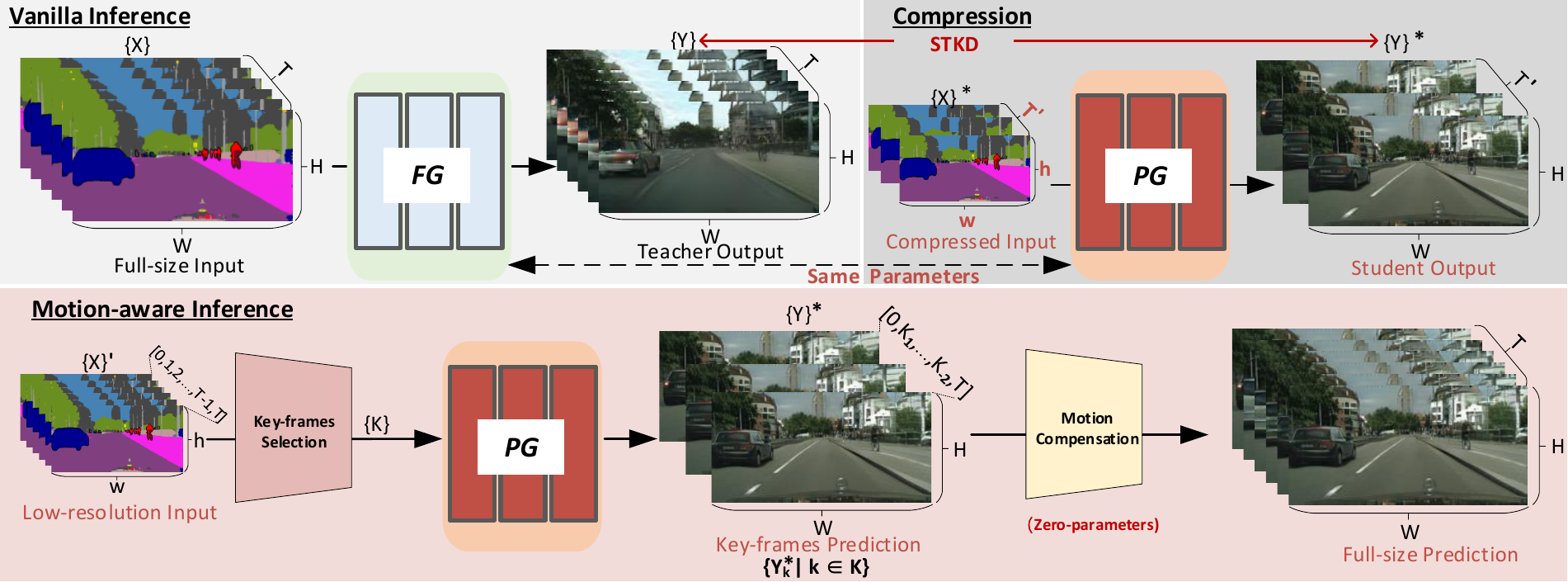}
  
  \caption{The pipeline of our Fast-Vid2Vid. It maintains the same amount of parameters as the original generator but compresses the input data in space and time dimensions. We perform spatial-temporal knowledge distillation (STKD) to transfer knowledge from the Full-time teacher Generator (FG) to the Part-time student Generator (PG). After STKD, Fast-Vid2Vid only infers the key-frames of the semantic sequence of low resolution and interpolates the intermediate frames by motion compensation.}
  \label{fig:infer}
\end{figure*}

To address the above issues, in this paper, we propose a novel spatial-temporal compression framework for vid2vid synthesis, named \textbf{Fast-Vid2Vid}. As shown in Fig.~\ref{fig:infer}, we reduce the computational resources by only compressing the input data stream through Motion-Aware Inference (MAI) without destroying the well-designed and complicated network parameters of the original Vid2Vid model, which addresses challenge 1. For challenge 2 and 3, we propose a Spatial-Temporal Knowledge Distillation method (STKD) that transfers spatial and temporal knowledge from the original model to the student network using compressed input data.
In particular, motivated by the spatial resolution-aware knowledge distillation method~\cite{feng2021resolution} that transfers the knowledge from large-size images to small-size ones for image recognition, our goal is to transfer the knowledge from large-size synthesized videos to small-size synthesized ones to make GAN robust enough to gain promising visual performance when the input data is compressed. 

We first train a spatially low-demand generator by taking low-resolution sequences as input but generating the full-resolution sequences. We perform Spatial Knowledge Distillation (Spatial KD) and transfer the spatial knowledge from the original generator to the spatially low-demand generator to obtain high-resolution frame information. Furthermore, we train a part-time generator by uniformly sampling video frames from sequences as real data. We perform Temporal-aware Knowledge Distillation (Temporal KD) and distill the temporal knowledge of the original generator to the part-time student generator to obtain full-time motion information by the introduced two losses, i.e., local temporal knowledge distillation loss and global temporal knowledge distillation loss. This design aims to capture the implicit knowledge in the time dimension.

To summarize, to the best of our knowledge, we make the first attempt to tackle the vid2vid compression problem at data aspects. On a single V100 GPU, Fast-Vid2Vid achieves \textcolor{orange}{18.56 FPS} (6.1$\times$ acceleration) with \textcolor{orange}{8.1$\times$} less computational cost on Sketch2Face, \textcolor{orange}{24.77 FPS} (5.8$\times$ acceleration) with \textcolor{orange}{8.3$\times$} less computational cost on Segmentation2City, and \textcolor{orange}{21.39 FPS} (7.1$\times$ acceleration) with \textcolor{orange}{9.3$\times$} less computational cost on Pose2Body. The main contributions of this paper are concluded as two-fold:

\begin{itemize}
    \item We present \textbf{Fast-Vid2Vid}, an sequential data stream compression method in spatial and temporal dimensions to greatly accelerate the vid2vid model. 
    \item We introduce a Spatial KD method that transfers knowledge from a teacher model input high-resolution data to a student model input low-resolution data to learn high-resolution information. 
    \item We propose a Temporal KD method to distill knowledge from a full-time teacher model to a part-time student model. A new temporal knowledge distillation loss globally is further presented to capture the time-series correlation. 
\end{itemize}

\section{Related Work}

\noindent\textbf{Video-to-Video Synthesis.} Video-to-video synthesis (Vid2vid) is a computer vision task that generates a photo-realistic sequence using the corresponding semantic sequence. Based on high-resolution image-based synthesis \cite{wang2018pix2pixHD}, Wang \textit{et al.}~\cite{wang2018video} developed a standard vid2vid synthesis model by introducing temporal coherence. Few-shot vid2vid model~\cite{wang2019few} further extended a few-shot version of the vid2vid model, which only uses fewer samples to achieve decent performance. Recently, vid2vid has been successfully extended to a wide range of video generation tasks, including video super-resolution~\cite{sajjadi2018frame,chu2020learning,xu2021temporal}, video inpainting~\cite{zou2021progressive,xu2019deep}, image-to-video synthesis~\cite{siarohin2019first,siarohin2021motion} and human pose-to-body synthesis~\cite{chan2019everybody,gafni2019vid2game,zhou2019dance,kappel2021high}. Most of these methods exploited temporal information to improve the performance of generated videos. However, they do not focus on vid2vid synthesis compression but on better visual performance.

\noindent\textbf{Model Compression.}
Model compression aims at reducing superfluous parameters of deep neural networks to accelerate inference. 
In the computer vision task, lots of model pruning approaches \cite{han2015learning,li2016pruning,hu2016data,liu2017learning,he2018soft,zhang2018systematic,wang2020gan} have greatly cut the weights of neural networks and significantly speed up inference time. Hu \textit{et al.}~\cite{hu2016network} reduced the unnecessary channels with low activations. Small incoming weights~\cite{he2018soft,li2016pruning} or outcoming weights~\cite{he2017channel} of convolution layers were used as saliency metrics for pruning. GAN compression has been proved by~\cite{yu2020self} that it is far more difficult than normal CNN compression. Due to the complex structures of GANs, a content-aware approach~\cite{liu2021content} was proposed to use salient regions to identify specific redundancy for GAN pruning. Wang \textit{et al.}~\cite{wang2020gan} reduced the redundant weights by NAS using a once-for-all scheme. 
Notably, the mentioned methods focus on simplifying the network structure and ignore the amount of the input information. Furthermore, these approaches do not consider the essential temporal coherence for video-based GAN compression, and thus achieve sub-optimal results for vid2vid models. Therefore, it is required to remove temporal redundancy in vid2vid models.

\noindent\textbf{Knowledge Distillation.}
Knowledge Distillation aims to make a student network imitate its teacher. Hinton \textit{et al.}~\cite{hinton2015distilling} proposed an effective framework for model distillation in classification. Knowledge distillation has been widely used in recognition\quad models~\cite{chen2017learning,chen2020distilling,lopez2015unifying,luo2016face,yim2017gift}. Recently, lots of response-based knowledge distillation methods~\cite{aguinaldo2019compressing,chen2020distilling,fu2020autogan,belousov2021mobilestylegan} were proposed for image-based GAN compression. For example, Jin \textit{et al.}~\cite{jin2021teachers} developed distillation techniques from ~\cite{li2020gan} and used a global kernel alignment module to gain more potential information. Liu \textit{et al.}~\cite{liu2021content} utilized a salient mask to guide the knowledge distillation process based on the norm. 
These methods only address image-based knowledge distillation, and thus only spatial knowledge is exploited, and they do not consider movements. It is not able to fully exploit temporal knowledge for vid2vid compression. Different from spatial-aware knowledge distillation, we consider both spatial information and temporal information into knowledge distillation, which tailors for vid2vid model compression. Most recently, Feng \textit{et al.}~\cite{feng2021resolution} have presented a resolution-aware knowledge distillation method that ignores the network parameters and compresses the input information for image recognition. In our work, we first introduce this input data compression method for GAN synthesis.

\section{Fast-Vid2Vid}

\subsection{A Revisit of GAN Compression}
\label{subsec:baseline}
 
The function of a deep neural network (DNN) can be written as $f(X) = W*X$, where $W$ denotes the parameters of the networks, $*$ represents the operation of DNN and $X$ denotes the input data. Obviously, two essential factors accounting for computational cost are the parameters and the input data. 
Existing GAN compression methods~\cite{aguinaldo2019compressing,chen2020distilling,fu2020autogan,jin2021teachers,liu2021content,li2020gan} have intended to cut the computational cost by reducing the parameters of network structures.
However, the network structures of GAN for specific tasks are carefully designed and it would cause poor visual results if the network parameters are cut arbitrarily. 
Another way to reduce computational cost is by compressing the input data. In this work, we seek for compressing the input data instead of the parameters of well-designed networks. To the best of our knowledge, there is little literature working on compressing data for GAN compression.

\subsection{Overview of Fast-Vid2Vid}
The typical Vid2vid framework~\cite{wang2018video} takes a sequence of semantic maps $\{X\}_{0}^T \in \mathbb{R}^{T\times H \times W }$ with $T$ frames and the initial real frames as input and predicts a photo-realistic video sequence $\{Y\}_{0}^T \in \mathbb{R}^{T\times H \times W}$. $H$ and $W$ denotes the height and weight of each frame. The vanilla inference of the vid2vid model (the full-time teacher generator) that utilizes a full-size sequential input data stream is a consecutive process that synthesizes a video sequence frame by frame. Considering both image synthesis and temporal coherence, a vid2vid model often contains several encoder-decoders to capture spatial-temporal cues, which are computationally prohibitive and even far from applications of mobile devices. In this paper, we propose a Fast-Vid2Vid compression framework, an input data compression method, to reduce the computational resources of the vid2vid framework in both space and time dimensions. 

Fig.~\ref{fig:infer} illustrates the overview of the proposed method. Fast-Vid2Vid first replaces the resBlock of the original vid2vid generator~\cite{wang2018video} with decomposed convolutional block~\cite{howard2019searching} to obtain a modern network architecture, which is similar to~\cite{li2020gan}. During knowledge distillation, we train a student generator using the compressed data and distill knowledge from the teacher generator by our proposed spatial-temporal knowledge distillation method (STKD). STKD, including spatial knowledge distilation (Spatial KD) and temporal knowledge distillation (Temporal KD), performs \textbf{spatial resolution compression} and \textbf{temporal sequential data compression}. After STKD, a part-time generator cooperating with motion compensation synthesizes a full-size sequence by \textbf{motion-aware inference (MAI)}.

\subsection{Spatial Resolution Compression for Vid2vid}
\label{subsec:spatial}
To reduce the spatial input data, a straightforward way~\cite{feng2021resolution} is to predict the low-resolution results using low-resolution semantic maps as the input sequence and re-size them into the full-resolution by a distortion algorithm. However, in our preliminary experiments, the straightforward method leads to severe artifacts since the distortion algorithm lacks high-frequency information and losses many important textures. Therefore, we make an adaptive change for vid2vid synthesis. We replace the downsampling layers with the normal convolution layers to generate the high-resolution results input by the low-resolution semantic maps. For formulation, the modified generator takes the low-resolution semantic sequence $\{X\}_{0}^{'T} \in \mathbb{R}^{T\times h \times w }$ as the input, where $h\times w= \frac{1}{(2^d)^2}H\times W$, and $d$ denotes the numbers of the modified downsampling layers. $d$ is set to 1. In this way, we obtain a spatially low-demand generator.

\begin{figure}[t]
\centering
\includegraphics[width=0.7\linewidth]{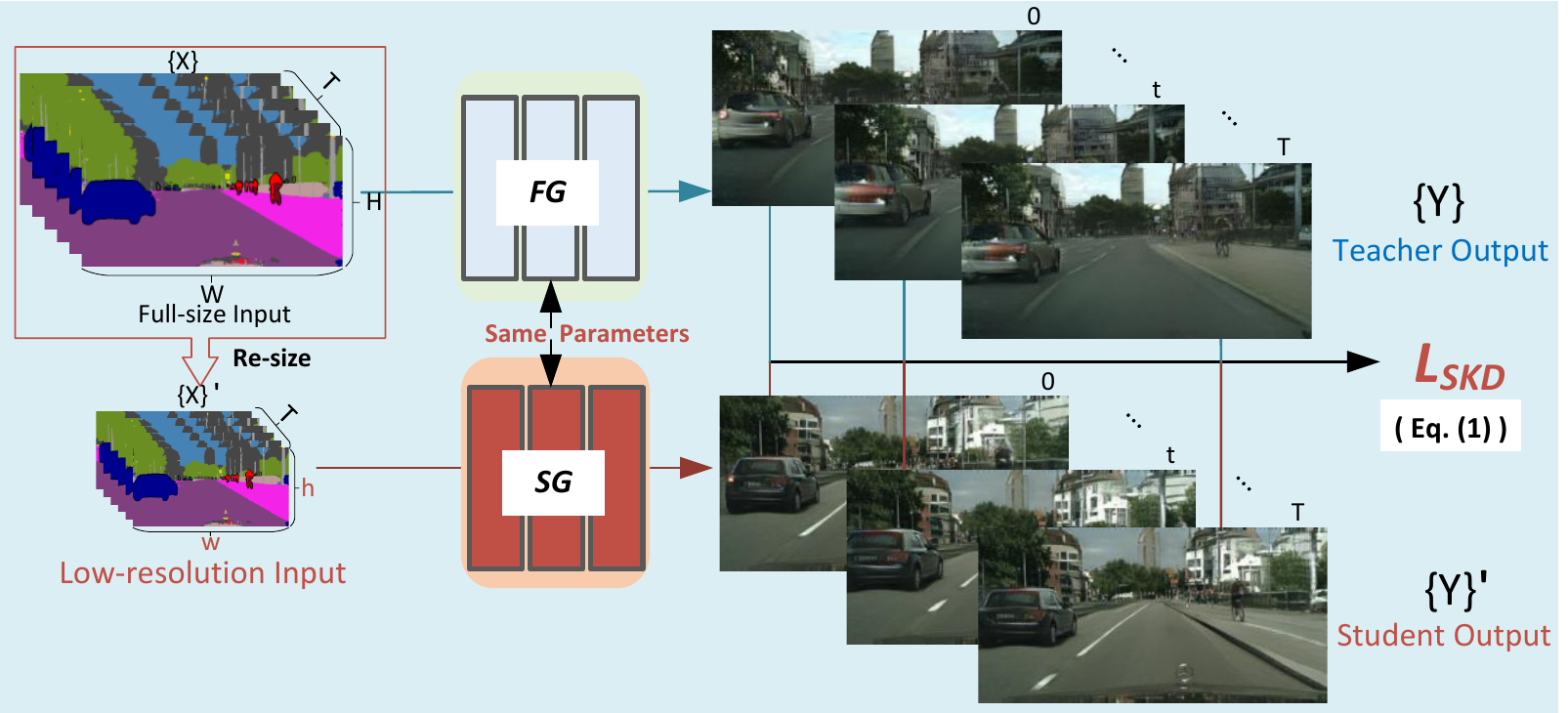}
\caption{The proposed Spatial Knowledge Distillation (Spatial KD). The spatially low-demand generator is fed with a sequence of low-resolution semantic maps and outputs full-resolution results. The results of the spatially low-demand generator are used for spatial knowledge distillation.}
\label{fig:skd_tkd}
\end{figure}

\begin{figure*}[t]
  \centering
  \includegraphics[width=\linewidth]{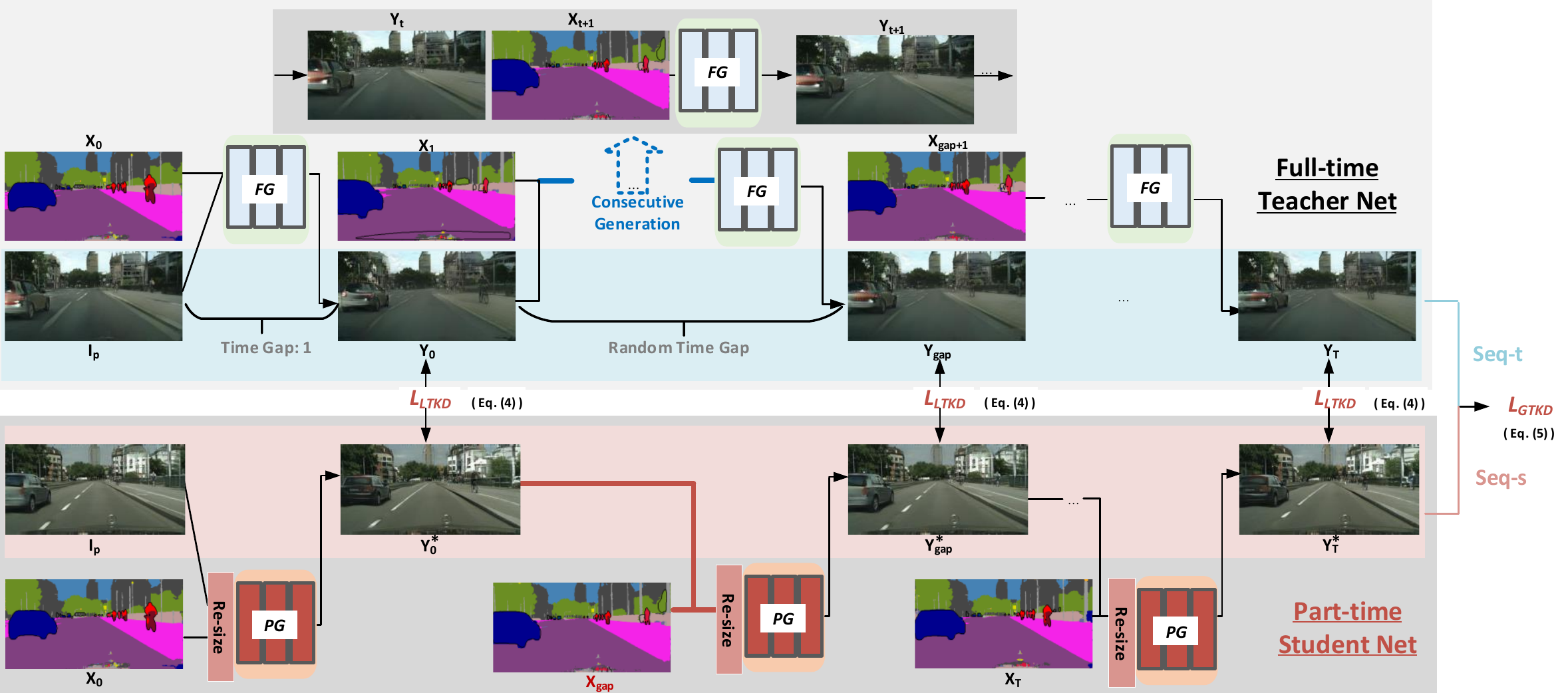}
  \caption{The proposed temporal-aware knowledge distillation (Temporal KD). The full-time generator and part-time generator synthesize the current frame using the previous frames and the semantic maps. The full-time teacher generator takes full-resolution semantic maps as inputs and generates a full sequence, while the part-time student generator takes only several low-resolution semantic maps as inputs and generates the corresponding frames at random intervals. }
  \label{fig:stkd}
\end{figure*}

Next, the spatially low-demand generator is required to learn high-frequency representation from the full-time teacher generator. We present a spatial knowledge distillation method (Spatial KD) to model high-frequency knowledge from the teacher net. Specifically, as shown in Fig.~\ref{fig:skd_tkd}, Spatial KD shrinks the margin between the low-resolution domain and the high-resolution domain to improve the performance of the student network. Spatial KD implicitly transfers spatial knowledge from the teacher net to the student net. Particularly, Spatial KD applies a knowledge distillation loss to mimic the visual features of the teacher net, and the loss function $L_{SKD}$ can be written as:
\begin{equation}
    L_{SKD} = \frac{1}{T}\sum_{t=0}^{T}[ MSE(Y_t,Y'_t)+ L_{per}(Y_t,Y{'}_t)],
\end{equation}
where $t$ means the current timestamp, $T$ is the total timestamps of the sequences, $L_{SKD}$ denotes the spatial knowledge distillation loss, $Y$ is the output sequence of the teacher net and $Y'$ is the predicted sequence of the spatially low-demand generator. $MSE$ represents a mean squared error between two frames. $L_{per}$ denotes a perceptual loss~\cite{wang2018video}.

\subsection{Temporal Sequential Data Compression for Vid2vid}
\label{subsec:temporal}

Each video sequence consists of dense video frames, which brings an enormous burden on computational devices. How to efficiently synthesize dense frames with a sequence of semantic maps is a difficult yet important issue for lightweight vid2vid models. 

In Section \ref{subsec:spatial}, we obtain a spatially low-demand generator. To ease the burden of generating dense frames for each video, we re-train the spatially low-demand generator on sparse video sequences, which are uniformly sampled from dense video sequences. The sampling interval is randomly selected in each training iteration. The original vid2vid generator is regarded as a full-time teacher generator and the re-trained spatially low-demand generator is regarded as a part-time student generator. To distill the temporal knowledge from the full-time teacher generator to the part-time student generator, we propose a local temporal-aware knowledge distillation method and a global temporal-aware knowledge distillation for temporal distillation, as shown in Fig.~\ref{fig:stkd}.

Both the full-time teacher generator and the part-time student generator take the previous $p-1$ synthesized frames $\{Y\}_{1}^{p-1}$ and $p$ semantic maps $\{X\}_{1}^p$ as input and generate the next frame. The previous frames are used to capture the temporal coherence of the sequences and generate more coherent video frames. The generation process of the full-time teacher generator is the consecutive generation. 
%Note that ${Y}_0$ is inferred by real frames and ${X}_0$. 
More generally, each generation iteration of the full-time teacher generator can be formulated as follows:
\begin{equation}
    Y_k = f_{FG}({\{X\}_{k-p}^{k}},{\{Y\}_{k-p}^{k-1}}),
\end{equation}
where $Y_k$ denotes the predicted current generative frame of the full-time teacher generator. $f_{FG}$ denotes the generation function of the full-time teacher generator. ${\{X\}_{k-p}^{k}}$ denotes $p+1$ frames of semantic maps, and ${\{Y\}_{k-p}^{k-1}}$ denotes the previous $p$ generated frames. 

Different from the full-time teacher generator whose uniform sampling interval is 1, the uniform sampling interval of the part-time student generator is $g$, where $1<g<T$. $g$ is a random number and randomly selected in each training iteration. Similarly, the frame generation of the part-time student generator can be formulated as follows:
\begin{equation}
    Y^{*}_k = f_{PG}(f_{R}^d({\{X_t\}_{\ k-p}^{*k}}),\{Y\}_{\ k-p}^{*k-1}),
\end{equation}
where $Y^{*}_k$ denotes the predicted current generative frame of the part-time student generator. $f_{PG}$ denotes the generation function of the part-time student generator, $f_{R}^d$ denotes the function of reducing the resolution into $\frac{1}{(2^d)^2}$, ${\{X\}_{\ k-p}^{*k}}$ includes $p$ frames of the sparse semantic sequences and $\{Y\}_{\ k-p}^{*k-1}$ is the previous frames of the synthesized sparse sequences.

To better illustrate our proposed Temporal KD, we set $p=1$ in Fig.~\ref{fig:stkd}. Specifically, the full-time teacher generator takes a semantic sequence $\{X\}_{0}^T$ as input and generates an entire sequence $\{Y\}_{0}^T$ frame by frame. For the $k$-th frame synthesis, the  full-time generator takes ${X_{k-1},X_k}$ and $Y_{k-1}$ as input and generates $Y_k$.

Because the full-time teacher generator is trained on sequences with dense frames and is learned to generate dense coherent frames, the full-time teacher generator cannot directly skip partial frames to generate sparse frames, leading to expensive computational cost. The part-time student generator can generate sparse frames and interpolate intermediate frames with the slight computational cost. However, since the part-time student generator is trained on sequences with sampled sparse frames, the low sample rate could be two times less than temporal motion frequency and thus leads to aliasing according to Nyquist–Shannon sampling theorem. Our preliminary experiments also show that the large changes between two non-adjacent frames cause remarkable inter-frame incoherence and generate a bad result. 

\noindent
\textbf{Local Temporal-aware Knowledge Distillation.}
We first introduce the local temporal-aware knowledge distillation to optimize the part-time student generator. Our goal is to distill the knowledge from the full-time generator to the part-time student generator to reduce aliasing. A straightforward idea is to align the outputs of the full-time generator and the outputs of the part-time student generator and reduce the distances between the corresponding synthesized frame pairs. The loss function of local temporal-aware knowledge distillation is given by
\begin{equation}
    L_{LTKD} = \frac{1}{T}\sum_{t=0}^{T}[ MSE(Y_t^{*},Y_t)+ L_{per}(Y_t^{*},Y_t)],
\end{equation}
where $L_{LTKD}$ denotes a local temporal-aware knowledge distillation loss, $Y'$ denotes the resulting frame of the spatially low-demand generator and $Y$ denotes the resulting frame of the teacher net. $MSE$ represents a mean squared error between two frames. $L_{per}$ denotes a perceptual loss~\cite{wang2018video}.

\noindent
\textbf{Global Temporal-aware Knowledge Distillation.}
Local temporal-aware knowledge distillation allows the part-time student generator to imitate the local motion of the full-time teacher generator. However, it does not consider the global semantic consistency. Therefore, we further propose a global temporal-aware knowledge distillation to distill global temporal coherence from the full-time generator to the part-time student generator. 

\begin{figure}[t]
\centering
\includegraphics[width=0.7\linewidth]{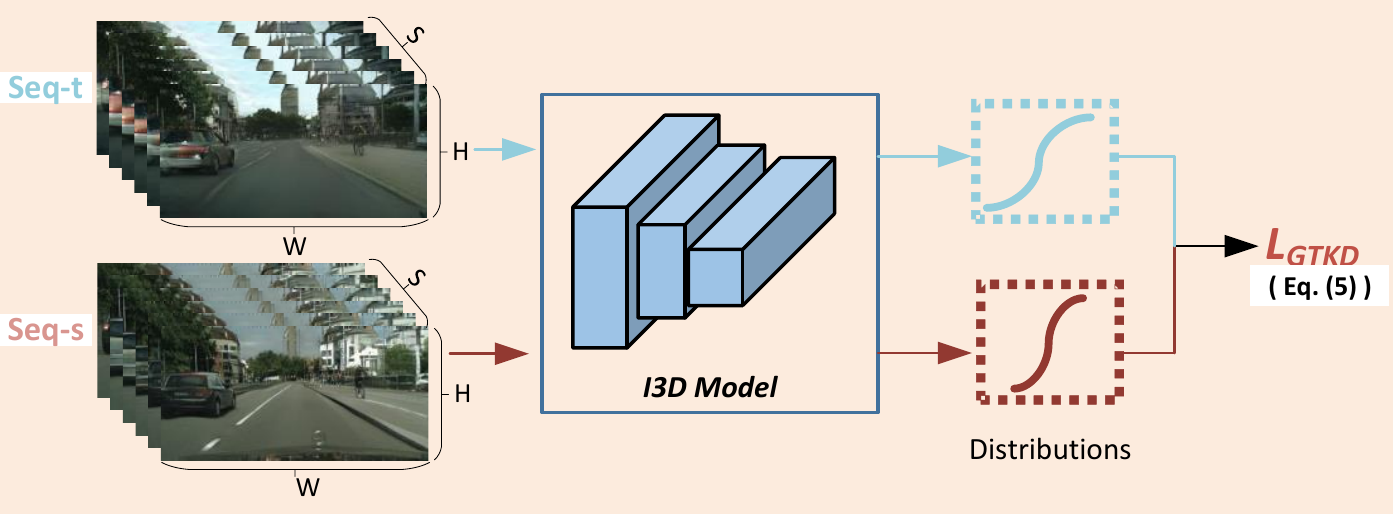}
\caption{The proposed temporal loss for temporal global knowledge distillation. The sequence of teacher net (seq-t) and the sequence of student net (seq-s) are extracted time-series coherence features by the well-trained I3D model for calculating the distances. }
\label{fig:skd_tkd2}
\end{figure}

It is observed that the current frame synthesis deeply relies on the results of the previous synthesis. This indicates that the generated current frame implicitly contains information from the previous frames. The global temporal-aware knowledge distillation exploits the generated non-adjacent frames generated by the full-time teacher generator to distill the hidden information of temporal coherence. The frames generated by the full-time teacher generator at the same timestamps as the non-adjacent frames by the part-time student generator are extracted and concatenated into a predicted sequence $\{Y^{*}\}$ (Seq-s). Similarly, the results of the full-time teacher generator are concatenated into a sequence $\{Y^{S}\}$ (Seq-t) in time order, as shown in the right figure of Fig.~\ref{fig:skd_tkd}. We introduce a global temporal loss to minimize the distance between $\{Y^{*}\}$ and $\{Y^{S}\}$, namely I3D-loss. The global temporal-aware knowledge distillation employs a well-trained I3D~\cite{carreira2017quo} model, a well-known video recognition model, to extract the time-series features of neighbor frames of $\{Y^{S}\}$ and $\{Y^{S}\}$. The global temporal-aware knowledge distillation loss is given by
\begin{equation}
    L_{GTKD} = MSE( f_{I3D}(\{Y^{*}\}\}), f_{I3D}(\{Y^{S}\})),
\end{equation}
where $MSE$ calculates the mean squared error between two feature vectors, $f_{I3D}$ is the function of the pre-trained I3D model. Finally, we obtain a temporal knowledge distillation loss, as written as,
\begin{equation}
    L_{TKD} = \alpha L_{LTKD} +\beta L_{GTKD},
\end{equation}
where $\alpha$ and $\beta$ control the importance of local and global losses. We set $\alpha=2$ and $\beta = 15$ to enhance the global temporal coherence.

\subsubsection{Full Objective Function.} 
We integrate local temporal-aware and global temporal-aware knowledge distillation into a unified optimization framework, which enables the part-time student generator to imitate both global and local motions of the full-time teacher generator. The full objective function is given by
\begin{equation}
    L_{KD} = \sigma L_{STD} +\gamma L_{TKD},
\end{equation}
where $\sigma$ and $\gamma$ control the weights of spatial and temporal KD losses, respectively. In particular, $\sigma$ is set as 1 and $\gamma$ is set as 2 to learn more knowledge of the temporal features from the teacher net.

\subsection{Semantic-driven Motion Compensation for Interpolation Synthesis}
\label{subsec:motion}
The temporal compression further greatly reduces the computational cost compared with the original vid2vid generator. However, the part-time student generator can only synthesize sparse frames ${Y'}$. To compensate for this problem, we use a fast motion compensation method \cite{orchard1994overlapped}, a zero-parameter algorithm, to complete the sequence. Motion compensation enables the synthesis of the inter frames between key-frames. As the adjacent frames are with slight changes, the final results remain a reliable visual performance by reducing the temporal redundancy. During inference, another question is which frames should be synthesized by the part-time student generator as sparse frames and how to determine these key-frames without sufficient photo-realistic frames. In this paper, we surprisingly find that we can distinguish key-frames $\{X'_k | k\in K\}$, where $K$ is a set containing the numbers of key-frame, from semantic maps $\{X\}_{0}^{'T}$. With the key semantic maps, the part-time student generator generates sparse frames $\{Y'_k| k \in K\}$ and finally, we interpolate other inter frames to a full-size result sequence $\{Y\} \in \mathbb{R}^{T \times H \times W}$ by the fast motion compensation method.

%%%%%%%%%%%%%%%%%%%%%%%%%%%%%%%%%%%%%%%%%%%%%%%%%%%%%%%%%%
\section{Experiments}

\subsection{Experimental Setup}

\noindent
\textbf{Models.}
We conduct our experiments using vid2vid\cite{wang2018video} model. The original vid2vid model uses a coarse-to-fine generator for high-resolution output. To simplify the compression problem, we only retrain the first-scale vid2vid model based on the official repository\footnote{https://github.com/NVIDIA/vid2vid}.
We also evaluate three compression methods for image synthesis models, including NAS compression~\cite{li2020gan}, CA compression~\cite{liu2021content} and CAT~\cite{jin2021teachers}. Since vid2vid is a 2D-based generation framework, these three methods can be easily transferred into this task with adjustments. For NAS compression, we adopt our full-time teacher generator to perform NAS with FVD the metric. For CA compression, we use salience-aware regions as the content of interest to compress the model. For CAT, the vid2vid model with modified residual blocks is built and retrained, followed by pruning and distillation using the global alignment kernel.

\noindent
\textbf{Datasets.}
%\footnote{We have provided the details of the datasets and more experimental results in supplementary materials.}
We evaluate our proposed compression method on several datasets. We pre-process the following datasets as the same as the settings of vid2vid, including Face Video dataset~\cite{rossler2018faceforensics}, Cityscape Video dataset~\cite{cordts2016cityscapes} and Youtube Dancing Video dataset~\cite{wang2018video}. Since we only apply the first scale of the vid2vid model, we re-size the datasets for convenience. Face Video dataset is re-sized into $512\times 512$, Cityscape Video dataset is re-size into $256 \times 512$, and Youtube Dancing Video dataset is re-sized to $384 \times 512$.

\noindent
\textbf{Evaluation Metrics.}
We apply three metrics for quantitative evaluation, including FID~\cite{heusel2017gans}, FVD~~\cite{unterthiner2018towards} and pose error~\cite{wang2019few}. FID~\cite{heusel2017gans} indicates the similarity between real and pseudo images using a well-trained classifier network. FVD aims to reveal the similarity between the real video and the synthesized video. Pose error measuring the absolute error in pixels between the estimated rendered poses and the original rendered poses predicted by Openpose~\cite{cao2017realtime} and Densepose~\cite{guler2018densepose}. The lower score of the three metrics represents better performance.

\noindent
\textbf{Key-frame Selection.}
We first calculate the residual maps between the adjacent frames, and sum up each map to draw smooth statistical curves using sliding windows. Thus, the peak ones of the curves represent the local maximum of the difference between two adjacent frames and are used as keyframes. Note that our keyframe selection only consumes about 0.5 milliseconds for processing a video of 30 frames.

\noindent
\textbf{Motion Compensation.}
Motion compensation is to predict a video frame given the previous frames and future frames in video compression, which is with fewer remnants than linear interpolation. We adopt an overlapped block motion compensation~(OBMC)~\cite{orchard1994overlapped} and an enhanced predictive zonal search~(EPZS) method~\cite{tourapis2002enhanced} to generate the non-keyframes by ``FFMPEG'' toolbox.  EPZS consumes about 2 MACs for each 16$\times$16 patch and OBMC consumes 5 MACs for each pixel, and thus requires 0.0008146G MACs for each video frame (512$\times$512 resolution), which is much less than our generative model part (282G MACs).

\begin{table*}[t]
\centering
\scriptsize
\caption{Quantitative results of Fast-Vid2Vid. m.MACs represents the mean of the total MACs of the calculation resources for a sequence of video.}
\label{tab:results}
\resizebox{0.8\textwidth}{!}{
\begin{tabular}{@{}c|ccc|ccc@{}}
\toprule
                                    &                                       &                                   &                                   & \multicolumn{3}{c}{Metric}                                                                                \\ \cmidrule(l){5-7} 
\multirow{-2}{*}{Task}              & \multirow{-2}{*}{Method      }              & \multirow{-2}{*}{m.MACs(G)}         & \multirow{-2}{*}{FPS}             & FID($\downarrow$)                & FVD   ($\downarrow$)              & PE($\downarrow$)              \\ \midrule
                                    & original                              & 2066                              &         2.77                      &    	34.17                        &    	6.74                           &   ---                             \\
                                    & NAS                                   &  303                            &         10.50                    &        33.64	                      &      6.86                       &    ---                            \\
                                    & CA                                    &    290                            &         11.07                     &    35.82                          &    6.95                           &    ---                               \\
                                    & CAT                                   &  294                              &        10.89                       &    33.90                          &    6.43                           &   ---                            \\
\multirow{-5}{*}{Sketch2Face}       & \textbf{Ours} & \textbf{282}  & \textbf{18.56} & \textbf{29.02} & \textbf{5.79} & --- \\ \midrule
                                    & original                              & 1254                              &   4.27                            &     ---                          &    2.76                           &  ---                              \\
                                    & NAS                                   & 277                               &   12.44                            &    ---                           &    2.99                           &  ---                             \\
                                    & CA                                    & 187                               &   15.48                           &    ---                           &   3.98                            &   ---                             \\
                                    & CAT                                   & 178                               &   13.29                            &    ---                            &    3.44                           &  ---                            \\
\multirow{-5}{*}{Segmentation2City} & \textbf{Ours} & \textbf{132} & \textbf{24.77} & --- & \textbf{2.33}& --- \\ \midrule
                                    & original                              &   1769                          &     3.01                         &   ---                         &      12.31                      &  2.60                            \\
                                    & NAS                                   &   280                            &   12.57                           &  ---                           &            12.53                     &   3.28                           \\
                                    & CA                                    &   253                              &   13.92                               & ---                              &   15.89                            &  4.85                           \\
                                    & CAT                                   & 257                               & 12.48                          &  ---                                         & 15.75              & 4.18       \\
\multirow{-5}{*}{Pose2Body}         & \textbf{Ours} & \textbf{191} & \textbf{21.39} & ---            & \textbf{10.03} & \textbf{2.18} \\ \bottomrule
\end{tabular}
%}
}
\end{table*}

\subsection{Quantitative Results}
We compare our compression method with the state-of-the-art GAN compression methods, NAS~\cite{li2020gan}, CA~\cite{liu2021content} and CAT~\cite{jin2021teachers} on three benchmark datasets to validate the effectiveness of our approach. For a fair comparison, we perform a decent pruning rate by removing around 60\% channels of the vid2vid model in CA and CAT, and use NAS to find out the best network configuration with 
similar mMACs.

The experimental results are shown in Table~\ref{tab:results}. 
We can see that given the lower computational budget, our method achieves the best FID and FVD on three datasets.  
Specifically, other GAN compression methods have lower performance than the full-size model while our method slightly outperforms the original model. Other compression methods speed up the original model by simply cutting the network structures, and they ignore the temporal coherence. Meanwhile, the original vid2vid model significantly accumulates losses during inference. Our proposed motion-aware inference accumulates less losses since it only generates several frames of the sequence. Such results show the advantage of our spatial-temporal aware compression methods.

\subsection{Ablation Study}

We adopt face video as the benchmark dataset for our ablation study.

\noindent\textbf{Effectiveness of Temporal KD Loss.}
Based on the spatially low-demand generator mentioned before, we analyze the knowledge distillation for the vid2vid model. We set 6 different distillation loss schemes as:
(1) w/o TKD: the spatially low-demand generator was retrained on the dataset;
(2) TKD-local: the spatially low-demand generator is transferred the only local knowledge from the teacher net;
(3) TKD-global: the spatially low-demand generator is transferred the only global knowledge from the teacher net;
(4) MMD: the spatially low-demand generator is transferred the knowledge using MMD-loss~\cite{feng2021resolution}.
(5) LSTM: the spatially low-demand generator is transferred the knowledge based on LSTM regulation~\cite{xiao2021space}.
(6) TKD: the spatially low-demand generator is transferred both local and global knowledge from the teacher net;

\begin{figure*}[t]
\centering
\subfigure[]{
\includegraphics[width=0.47\linewidth]{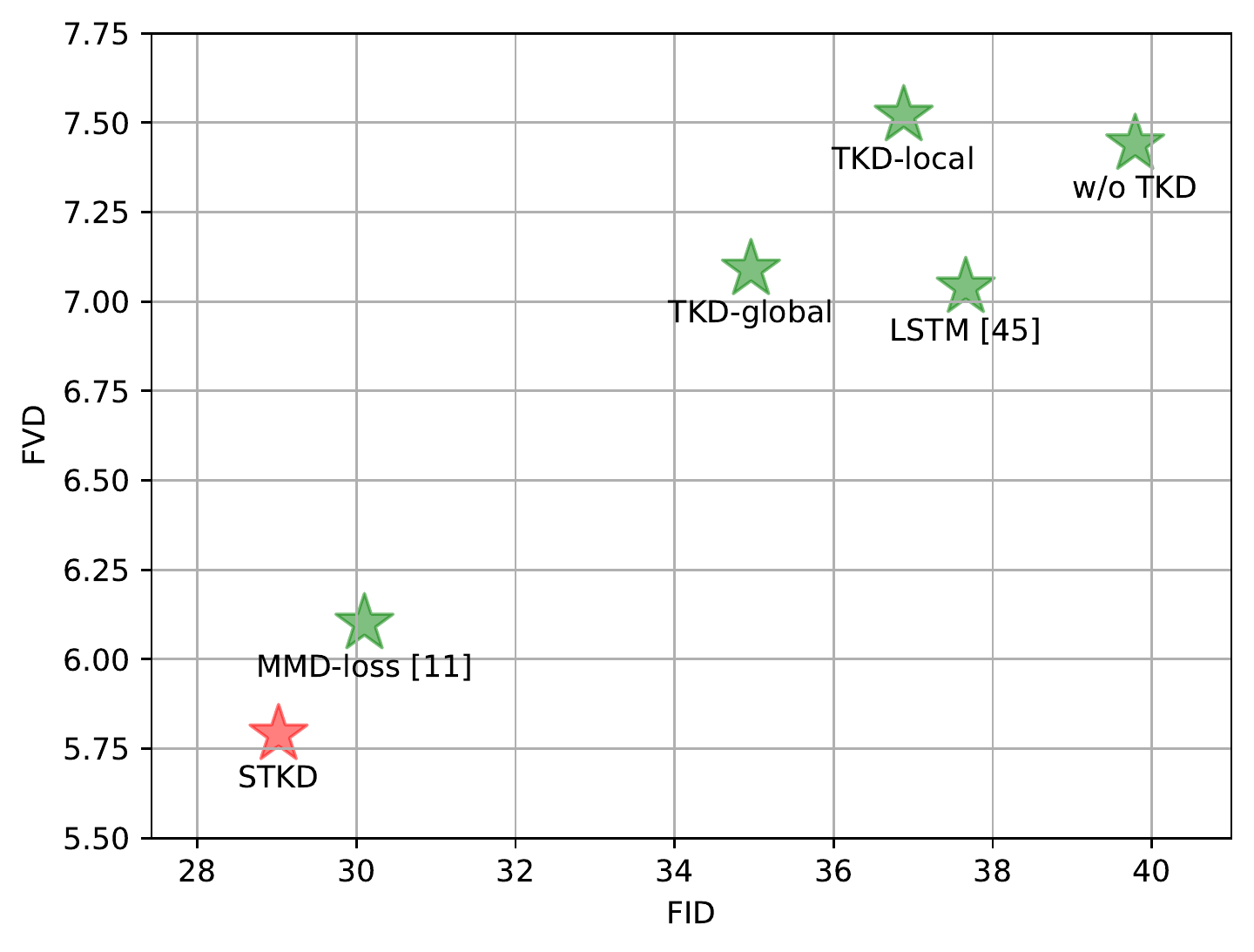}
}
\quad
\subfigure[]{
\includegraphics[width=0.45\linewidth]{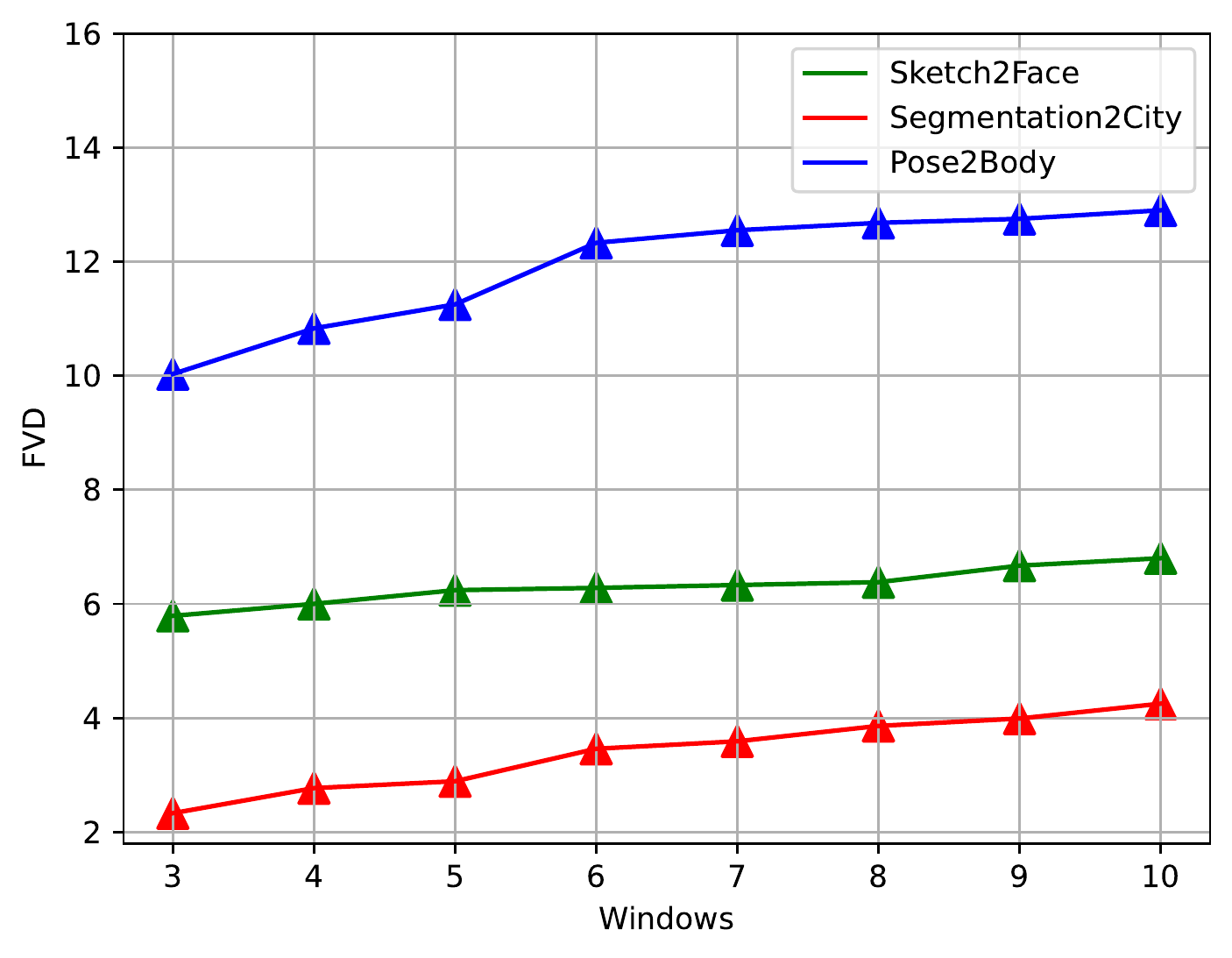}
}

\caption{Ablation Study for Fast Vid2Vid. (a) Temporal Loss ablation study for temporal loss. (b) The trade-off experiments for the windows of key-frames selector. Larger windows means less mMACs.}
\label{fig:ablation-loss-table}

\end{figure*}

\begin{figure*}[t]
\centering
\includegraphics[width=1\linewidth]{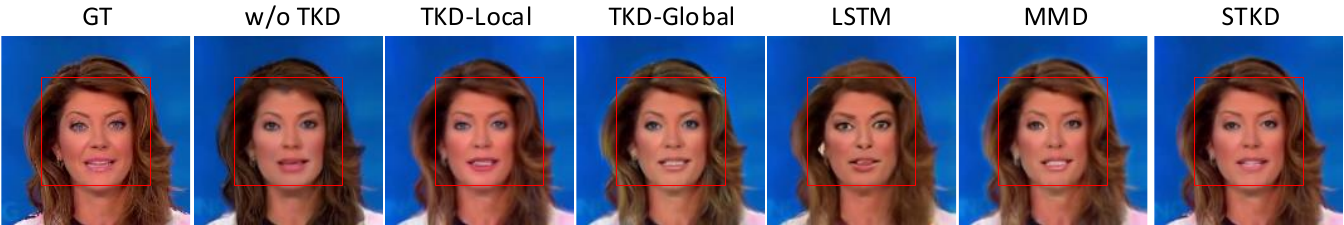}
\caption{The comparison of the results of different knowledge distillation.}
\label{fig:ablation-loss}
\end{figure*}

As shown in Fig.~\ref{fig:ablation-loss-table}(a), the local knowledge distillation loss improves the performance of the model without KD. Furthermore, the temporal KD loss globally further improves the performance of the common local KD loss, especially in FVD. Moreover, our proposed KD loss outperforms MMD-loss and LSTM-based KD loss. It indicates that the temporal KD loss effectively enhances the similarity of distribution on the video recognition network between the videos generated by the teacher network and the student network. 
We also provide the qualitative comparison among these KD methods in Fig.~\ref{fig:ablation-loss} and our STKD generates more photo-realistic frames than others.

\noindent\textbf{Effectiveness of Spatial KD Loss.}
We conduct an ablative study for Spatial KD on the Sketch2Face benchmark. In the {{video setting}}, spatial compression methods are used together with our proposed Temporal KD to perform vid2vid compression.  
Table \ref{abl} shows that our proposed Spatial KD performs better than other image compression methods. 
Our Spatial KD does not destroy network structures of the original GAN while other methods tweak the sophisticated parameters.

\begin{table}[t]
\centering
\caption{Ablation Study for spatial compression with the proposed Temporal KD. } 
\label{abl}
\resizebox{0.5\linewidth}{!}{
\begin{tabular}{@{}ccccc@{}}
\toprule
Method                   & MACs(G) & FPS   & FID   & FVD  \\ \midrule
CA                &    {331}    & {17.00}      &  {36.65}    &  {6.76}  \\
CAT                &  {310}      & {18.02}     & {35.64}     &   {6.85}  \\
NAS                &    {344}  &  {16.78}     & {32.41}  &   {6.71}   \\
Spatial KD  & \textbf{{282}}      & \textbf{18.56 }  & \textbf{{29.02}} & \textbf{{5.79}}  \\ \bottomrule
\end{tabular}
}
\end{table}

\noindent\textbf{Impact of Windows for Key-frame selection.} We investigate the sliding windows to select the key-frames. The larger sliding windows mean that there are fewer key-frames selected and thus use less computational resources. We aim to find out the best trade-off between the sliding windows and the performance. Interestingly, as shown in Fig. \ref{fig:ablation-loss-table}(b), it shows a significant rise in FVD when increasing the sliding windows, and achieves the best performance when the sliding windows are three in three tasks. It indicates that the part-time student generator needs enough independent motion to maintain decent performance.

\begin{table}[t]
\centering
\scriptsize
\makebox[0pt][c]{\parbox{1.0\textwidth}{%
    \begin{minipage}[b]{0.48\hsize}\centering 
                \caption{The performance of different interpolation.}
                \label{tab:inter}
        \begin{tabular}{ccc}
        \toprule
        Interpolation & FID            & FVD           \\
        \midrule
        Linear Interpolation    & 31.55          & 6.45         \\\midrule
        Motion Compensation  &  \textbf{29.02} & \textbf{5.79} \\
        \bottomrule
        \end{tabular}

            \label{tab:singlebest}
        \end{minipage}
        \hfill
        \begin{minipage}[b]{0.48\hsize}\centering
        
        \caption{The performance of different time gap.}
        \label{tab:tg}
            \begin{tabular}{ccc}
            \toprule
            Gap & FID            & FVD           \\
            \midrule
            Fixed time gap    &   32.21      &   6.85 \\ \midrule
            Random time gap & 31.43&6.22\\ \midrule
            Key-frames  &  \textbf{29.02} & \textbf{5.79} \\
            \bottomrule
            \end{tabular}
        \label{tab:threebest}
    \end{minipage}%
}}
\vspace{-18pt}
\end{table}

\noindent\textbf{Effectiveness of Interpolation.} We conduct two common interpolation methods for completing the video, namely linear interpolation and motion compensation. We also conduct ablative studies on interpolation methods. As we can see in Table~\ref{tab:inter}, motion compensation outperforms the simple linear interpolation. Therefore, we apply motion compensation as our zero-parameters interpolation method.

\noindent\textbf{Effectiveness of Time Gap.} We also study the ways of selected frames for the generation of the part-time student generator. We use the same numbers of the selected frames. Specifically, the fixed time gap strategy selects the frames in fixed time intervals, the random time gap strategy chooses the frames in random intervals, and the key-frame strategy selects the key-frames as the frames to be synthesized by the part-time student generator. As shown in Table~\ref{tab:tg}, the key-frame strategy outperforms other strategies since the key-frames of a sequence consist of all essential motions and texture.

\begin{figure*}[t]
\centering
%\subfigure[Sketch2Face]{
\includegraphics[width=1\linewidth]{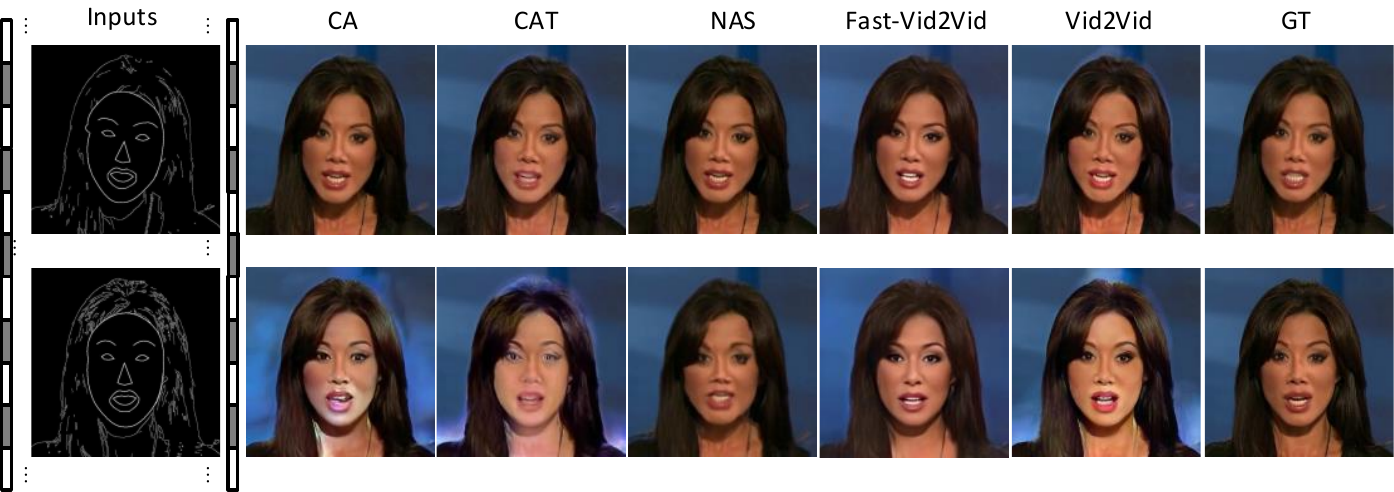}
%}
%\quad
%\subfigure[Segmentation2City]{
\includegraphics[width=1\linewidth]{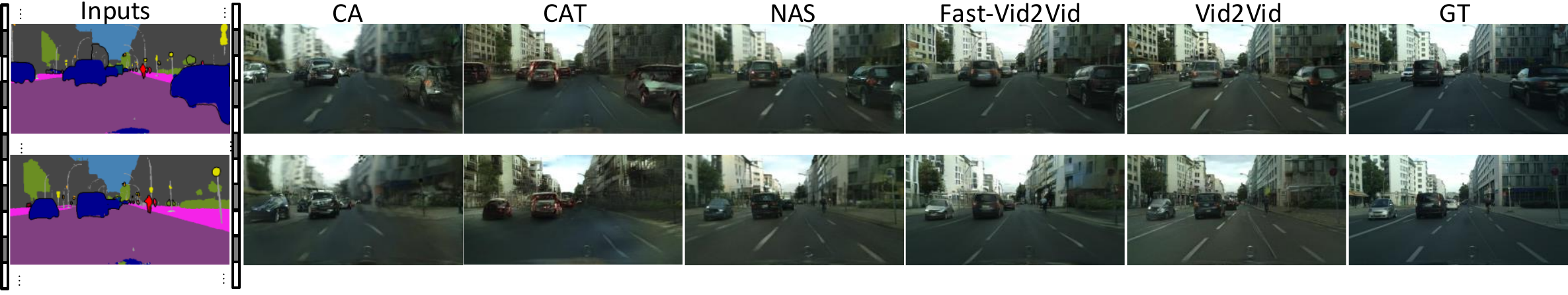}
%}
%\quad
%\subfigure[Pose2Body]{
\includegraphics[width=1\linewidth]{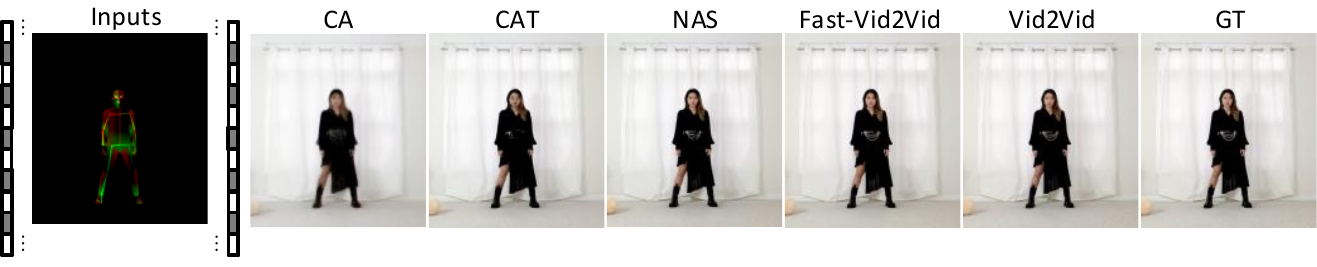}
%}
% \subfigure[The trade-off for the windows of key-frames selector.]{
% \includegraphics[width=0.47\linewidth]{Figures/line.pdf}
% }
\caption{Qualitative results compared with the advanced GAN compression methods in the task of Sketch2Face, Segmentation2City and Pose2Body. }
\label{fig:vis}
\end{figure*}

\subsection{Qualitative Results}

We illustrate the output sequences of the mentioned methods in Fig.~\ref{fig:vis}. The generators in vid2vid synthesis would accumulate the visual losses sequentially since the current frame relies on the previously generated frames. We also visualize more examples in Fig.~\ref{fig:vis2}, Fig.~\ref{fig:vis3} and Fig.~\ref{fig:vis4}. Compared with other GAN compression methods, our proposed method generates more realistic results.

\section{Discussion}
We discuss some future directions for this work. Recently, sequence-in and sequence-out methods, like transformer, are challenging for model compression. On the contrary, our Fast-Vid2Vid accelerates Vid2Vid by optimizing a part-time student generator (via temporal-aware KD compression) and a lower-resolution spatial generator (via spatial KD compression), which is versatile for various networks. When combined with seq-in and seq-out transformers like visTR~\cite{wang2021end}, Fast-Vid2Vid first synthesizes a partial video by a part-time transformer-based student generator (via fully parallel computation) and then recovers the full video by motion compensation.

\section{Conclusion}
In this paper, we present Fast-Vid2Vid to accelerate vid2vid synthesis. We propose a spatial-temporal compression framework to accelerate the inference by compressing the sequential input data stream but maintaining the parameters of the network. In space dimension, we distill knowledge from the full-resolution domain to the low-resolution domain and obtain a spatially low-demand generator. In time dimension, we use temporal-aware knowledge distillation for local and global knowledge to convert the spatially low-demand generator from a full-time generator to a part-time generator. Finally, the part-time generator is used for motion-aware inference where it only generates the key-frames of the sequence and interpolates the middle frames by motion compensation. By reducing the resolution of the input data and extracting the key-frames of the data stream, Fast-Vid2Vid saves the computational resources significantly.

\section*{Acknowledgements}
This work is supported by NTU NAP, MOE AcRF Tier 1 (2021-T1-001-088), and under the RIE2020 Industry Alignment Fund – Industry Collaboration Projects (IAF-ICP) Funding Initiative, as well as cash and in-kind contribution from the industry partner(s).

\bibliographystyle{splncs04}
\bibliography{egbib}

\appendix

\label{sec:appendix}

\begin{figure*}[t]
\centering

\includegraphics[width=1\linewidth]{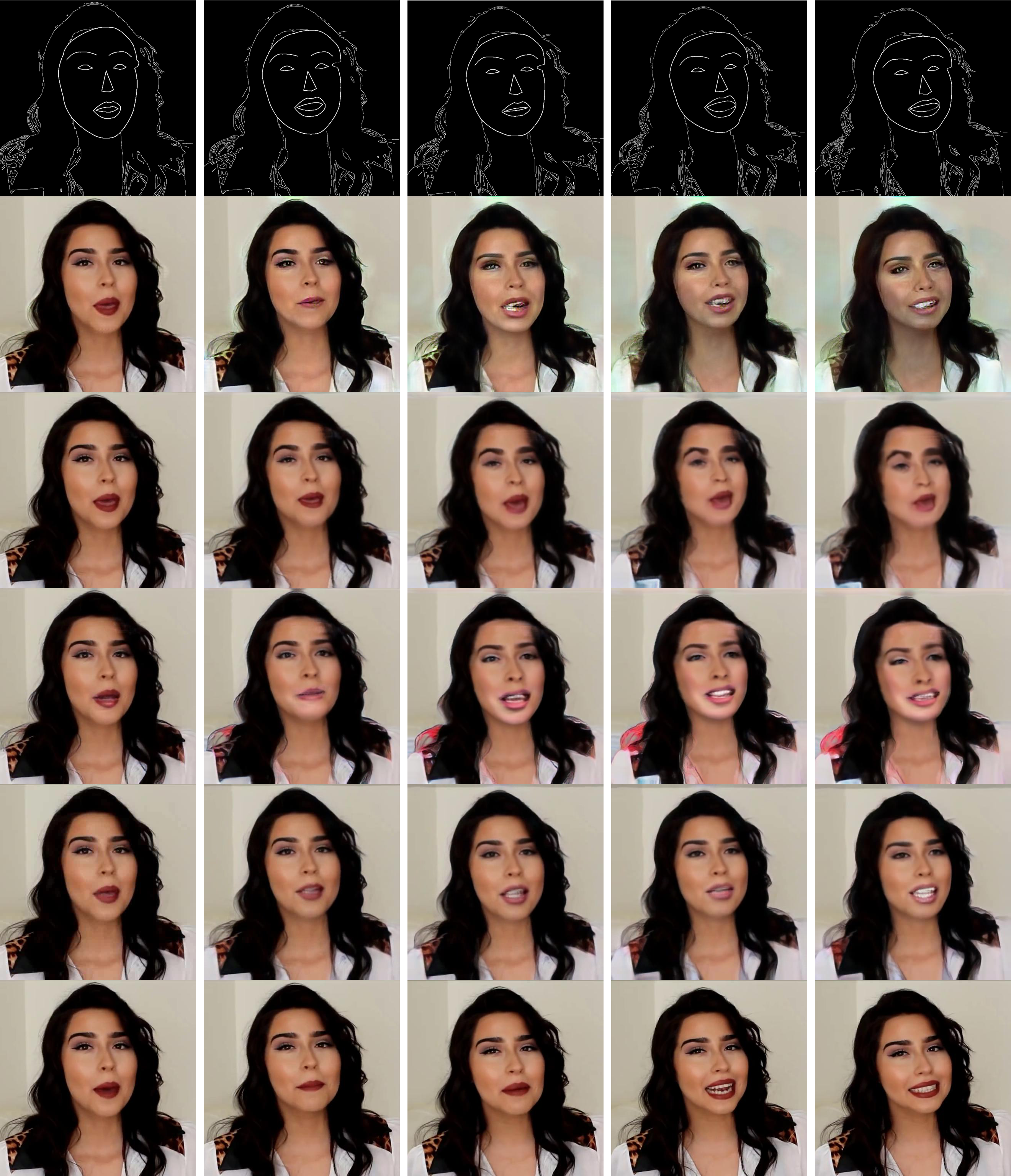}

\caption{Qualitative results of the testing data compared with the advanced GAN compression methods in the task of Sketch2Face. From top to the bottom, rows are semantic maps, CA's results, CAT's results, NAS's results, Vid2Vid's results, Fast-Vid2Vid's results and GT.}
\label{fig:vis2}
\end{figure*}

\begin{figure*}[t]
\centering

\includegraphics[width=1\linewidth]{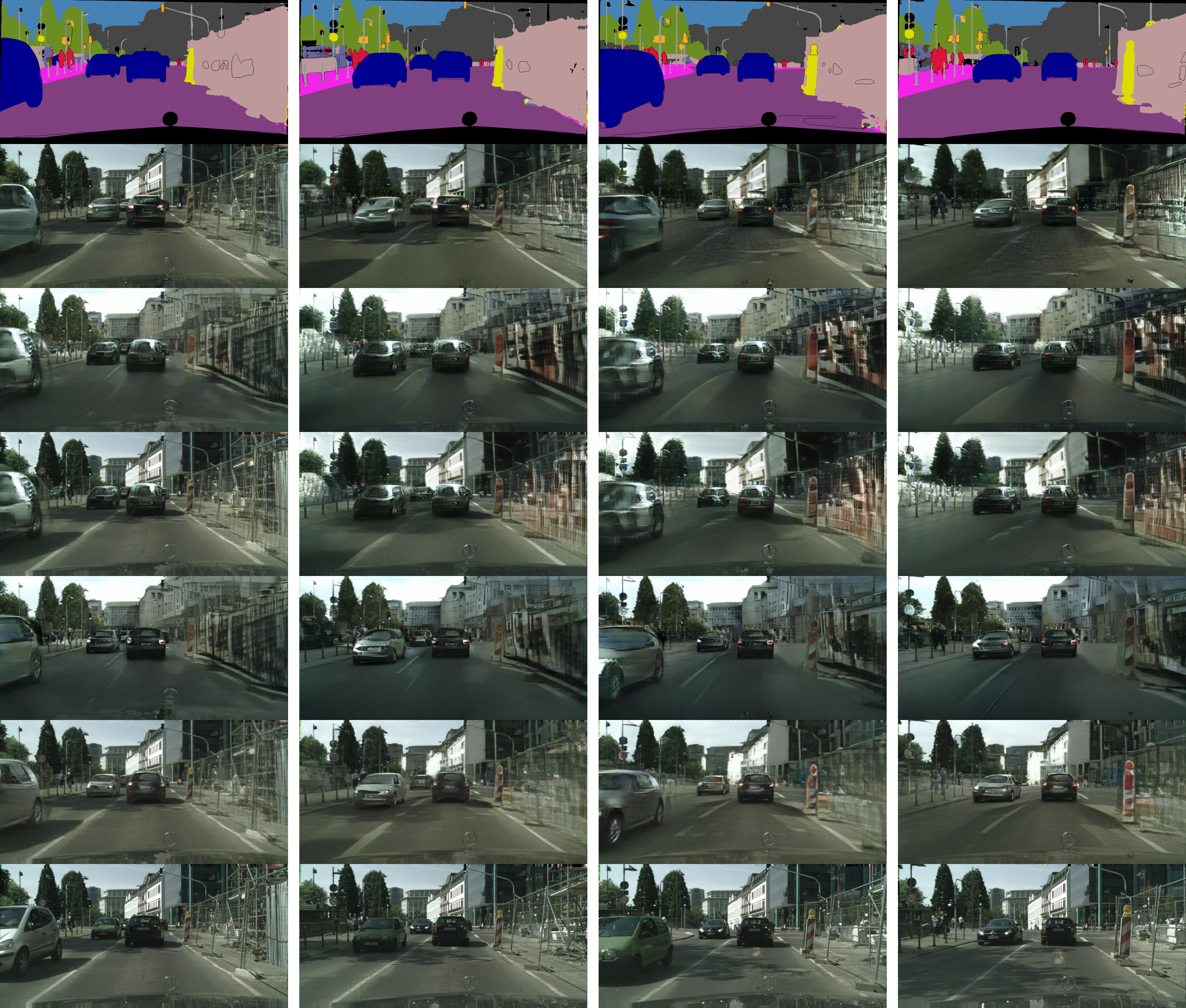}

\caption{Qualitative results of the testing data compared with the advanced GAN compression methods in the task of Segmentation2City. From top to the bottom, rows are semantic maps, CA's results, CAT's results, NAS's results, Vid2Vid's results, Fast-Vid2Vid's results and GT.}
\label{fig:vis3}
\end{figure*}

\begin{figure*}[t]
\centering

\includegraphics[width=1\linewidth]{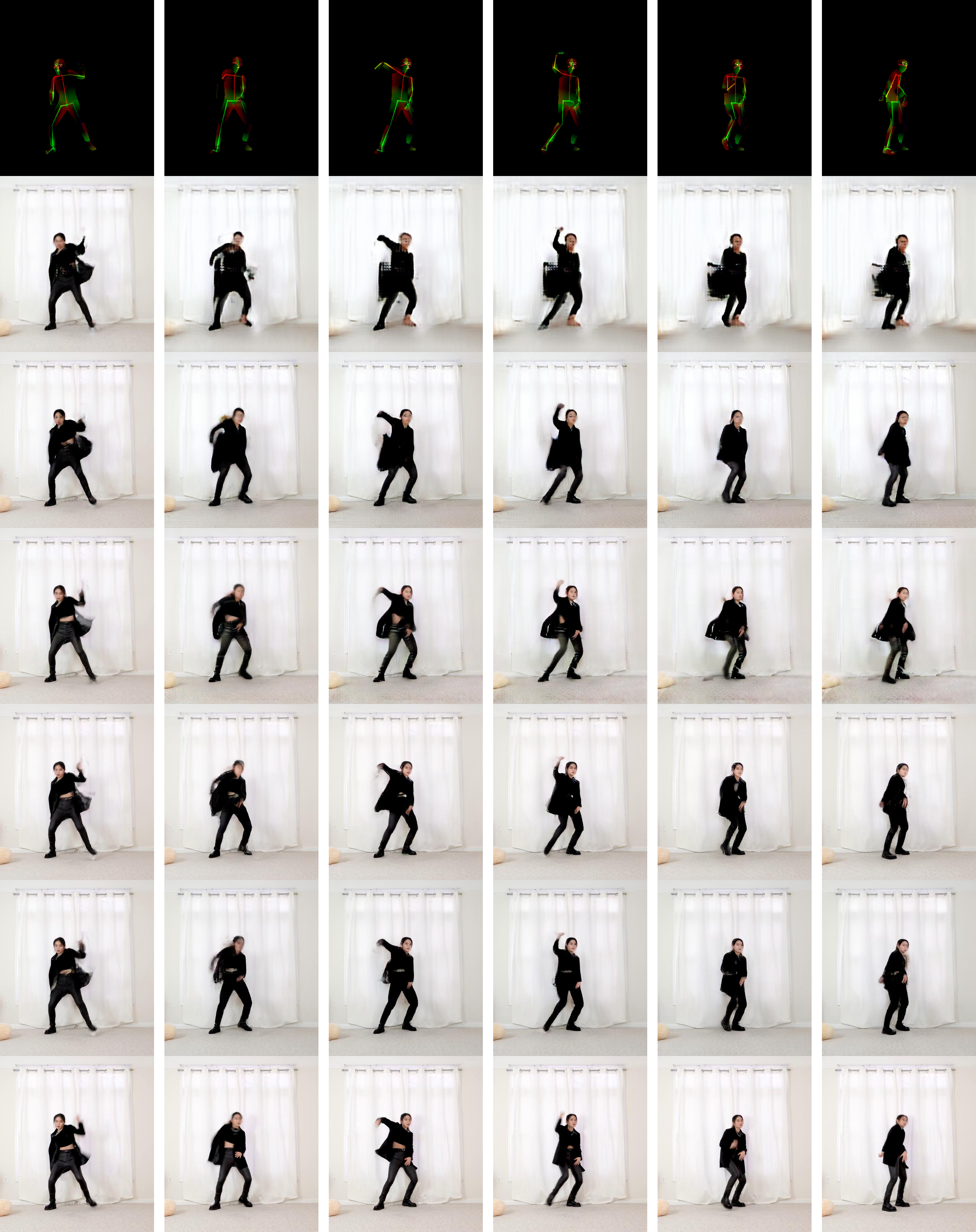}

\caption{Qualitative results of the testing data compared with the advanced GAN compression methods in the task of Pose2Body. From top to the bottom, rows are semantic maps, CA's results, CAT's results, NAS's results, Vid2Vid's results, Fast-Vid2Vid's results and GT.}
\label{fig:vis4}
\end{figure*}

\end{document}